\documentclass[11pt,a4paper,reqno, dvipsnames]{article}

\usepackage{PRIMEarxiv}

\usepackage[utf8]{inputenc} 
\usepackage[T1]{fontenc}    
\usepackage{hyperref}       
\usepackage{url}            
\usepackage{booktabs}       
\usepackage{amsfonts}       
\usepackage{nicefrac}       
\usepackage{microtype}      
\usepackage{fancyhdr}       
\usepackage{graphicx}       
\usepackage{mathtools, amsmath, amsthm, amsfonts, amssymb} 
\usepackage{bm}
\usepackage{cleveref}
\crefname{section}{§\hspace{-0.1cm}}{§§}
\Crefname{section}{§}{§§}
\usepackage{caption}
\usepackage{subcaption}
\usepackage{tikz} 
\usetikzlibrary{arrows.meta} 
\usetikzlibrary{positioning, shapes.multipart, arrows}
\usetikzlibrary{shapes, fit}
\usepackage{pgfplots}
\pgfplotsset{compat=1.18}
\usepackage{comment} 

  {\list{}{\leftmargin=#1\rightmargin=#1}\item[]}%
  {\endlist}

\usepackage{xspace}

\usepackage{epigraph}

\usepackage{todonotes}
\usepackage[pagewise]{lineno}

\usepackage{natbib}
\setcitestyle{aysep={}}
\usepackage[toc,page]{appendix} 

\title{a case for ai consciousness: 
language agents and global workspace theory

}

\author{
Simon Goldstein \\
  The University of Hong Kong \\
  \texttt{simon.d.goldstein@gmail.com} \\
  \And
  Cameron Domenico Kirk-Giannini \\
  Rutgers University-Newark \\
  \texttt{camerondomenico.kirkgiannini@gmail.com}\\ 
}

\begin{document}
\maketitle

\begin{abstract}
It is generally assumed that existing artificial systems are not phenomenally conscious, and that the construction of phenomenally conscious artificial systems would require significant technological progress if it is possible at all. We challenge this assumption by arguing that if Global Workspace Theory (GWT) — a leading scientific theory of phenomenal consciousness — is correct, then instances of one widely implemented AI architecture, the \emph{artificial language agent}, might easily be made phenomenally conscious if they are not already. Along the way, we articulate an explicit methodology for thinking about how to apply scientific theories of consciousness to artificial systems and employ this methodology to arrive at a set of necessary and sufficient conditions for phenomenal consciousness according to GWT.
\end{abstract}

The advent of generative text and image models has led to a resurgence of academic and popular interest in the question of what it would take for an AI to be conscious. While it is difficult to draw general conclusions on this topic because of the great diversity of philosophical and scientific theories of consciousness, a number of authors have recently suggested that it may be possible to construct conscious AIs in the near future (\citet{butlin2023, chalmers2023}). Recent work has focused on supporting this claim by surveying the commitments of a wide range of philosophical and scientific theories of consciousness. While this approach has the advantage of appealing to diverse theoretical perspectives, however, it is constrained in its level of detail. In what follows, we take a different approach. We focus on one leading scientific theory of consciousness, \emph{global workspace theory} (GWT), and one widely implemented AI architecture, the \emph{artificial language agent} (or \emph{language agent} for short), arguing that if GWT is correct, then language agents might easily be made conscious if they are not conscious already. Our project makes three main contributions to the literature on consciousness in artificial systems. First, it provides an explicit methodology for thinking about how to apply scientific theories of consciousness to artificial systems. Second, it focuses scholarly attention on language agents, which we take to be an especially interesting case study in artificial consciousness. Third, it presents a case for consciousness in AI systems that is better equipped to respond to certain important objections than earlier proposals.\footnote{ Since the 1990s, there have been a number of  attempts to implement a global workspace in an AI system. Notable  examples include CMattie \citep{franklin1999}, a simple system designed to schedule seminars at a university by sending and reading messages in natural language, its successor system IDA \citep{franklin2003}, and the robotic system designed by \citet{shanahan2006}. Though we think this existing work is interesting, its significance is limited by two factors. First, authors in this tradition  do not generally develop and defend an explicit account of the necessary and sufficient conditions for consciousness according to GWT. This makes it difficult to assess whether the systems claimed to implement a global workspace in fact do. Second, the systems described in this tradition are generally very simple and have a limited range of capabilities which could not be extended without significant architectural modifications, leaving them more vulnerable than language agent architectures to the ``small model objection'' (see section 9 for discussion).}

To investigate whether AI systems can be conscious, we adopt a broadly computational and functionalist perspective according to which consciousness is a special kind of information processing. AI systems are conscious if they process information in this special way. Within this framework, the most important open problem is to specify carefully which functional roles are associated with consciousness. Below, we describe how existing work on GWT has vacillated between a series of different functional conditions associated with consciousness. These conditions differ in how demanding they are and in their level of abstraction. We show that these differences matter to predictions about what it would take for AI systems to be conscious. We also develop two methodological tools to adjudicate between these various functional conditions. In particular, we reflect on the usefulness of consciousness and explore thought experiments that probe how our judgments about consciousness vary with slight changes to functional role. In each case, we argue that our methodological tools tend to favor more permissive functional roles, which make it easier for a wide range of systems to count as conscious.

Here is the plan. In section 1, we distinguish between access consciousness and phenomenal consciousness and discuss the place of each in the science of consciousness and normative theory. In section 2, we introduce GWT and describe some of the evidence for it. Sections 3, 4, and 5 distill from GWT a functional characterization of consciousness in artificial systems. Section 6 turns to contemporary machine learning, introducing the architecture of language agents. In section 7, we argue that existing language agents come close to satisfying the functional criteria introduced in section 5, and that they could be made to do so fully with simple architectural modifications. In section 8, we consider a number of further modifications to our proposed architecture that could address a range of different kinds of skepticism about its plausibility as a conscious system. Section 9 considers two objections to our arguments. Section 10 concludes.

\section{Consciousness}

The ordinary concept of a conscious state or being is imprecise, failing to differentiate among a diverse set of properties. For this reason, it is customary to distinguish between two important senses in which a state or being can be conscious: access consciousness and phenomenal consciousness. A state is access conscious to the extent that it is available for use in reasoning and guiding speech and action, while it is phenomenally conscious just in case it has experiential properties \citep{block02}. Correspondingly, a being is conscious in the access sense to the extent that it tokens access-conscious states and conscious in the phenomenal sense just in case it tokens phenomenally conscious states. 

While there are interesting scientific questions to ask about the kinds of biological and artificial functional architectures which give rise to access-conscious states, it is phenomenal consciousness which is generally regarded as the greater puzzle both scientifically and philosophically. Substantial literatures exist in philosophy, psychology, and cognitive neuroscience on the nature of phenomenal consciousness and its relation to physical systems.\footnote{For an introduction with suggestions for further reading, see \citet{van-gulick2022}.} In what follows, our primary concern will be with phenomenal consciousness. We will discuss one leading scientific theory of phenomenal consciousness in biological systems — Global Workspace Theory — and distill from it an account of the conditions under which artificial systems might also be phenomenally conscious. We take it to be uncontroversial that any artificial system meeting these conditions would be access conscious, and we will not argue at length for this claim.

The question of consciousness in artificial systems is especially urgent because many philosophers tie questions of consciousness to the normative domain. First, it is commonly held that certain phenomenally conscious states contribute positively or negatively to an individual’s wellbeing in virtue of their phenomenal properties. For example, the pleasantness of pleasures and the unpleasantness of pains are held to improve and reduce an individual’s wellbeing, respectively.\footnote{See, for example, \citet{kagan1992} and \citet{bramble2016}.} This connection pertains to phenomenal consciousness rather than access consciousness. Second, it is often held that a being must be conscious in order for it to be the right kind of thing to have wellbeing in the morally significant sense.\footnote{See \citet{lin2021} and \citet{goldstein2023} for recent discussion.} According to this line of thought, while it might be said of simple systems like protists and fungi that things could go well or poorly for them, it does not follow that we have a moral reason to avoid harming them because they are not conscious.

\section{Global Workspace Theory}

Global Workspace Theory originated with the work of Bernard Baars \citep{baars1988, baars1997b} and has since grown to be one of the leading theories of consciousness in biological systems. GWT posits the existence of a \emph{global workspace}: a component of cognitive architecture that receives information from a range of cognitive modules, processes this information, and then broadcasts the processed information back to the cognitive modules to which it is connected. According to GWT, a representation is conscious when it is present in the global workspace.\footnote{This is a slight simplification. \citet{baars1988} (363), for example, suggests that representations may need to be present in the global workspace for some minimum amount of time before they are consciously experienced, and \citet{baars1997a} (44) suggests that representations in the global workspace must be in ``the spotlight of attention'' to be consciously experienced. We set these complications aside in what follows since we do not take them to bear significantly on our arguments.} Though proponents of GWT are not always clear on the sort of consciousness they have in mind, we interpret GWT as endorsing the claim that information in the global workspace is conscious in both the access and phenomenal senses, and that being in the global workspace is what renders a piece of information phenomenally conscious.

It is helpful to distinguish the various claims associated with GWT into \emph{structural} claims about cognitive architecture and \emph{functional} claims about how the global workspace receives information, processes it, and transmits it to other parts of the cognitive system. The core structural claim of GWT is that human cognitive architecture consists of a number of relatively autonomous \emph{modules} which process information specific to particular tasks (e.g. particular sense modalities, motor control, and so on) together with a single global workspace with which all of these modules interface.\footnote{For example, \citet{dehaene1998} mention five main modules connected to the global workspace: perceptual systems, long-term memory, evaluative systems, attentional systems, and motor systems (see also \citet{mesulam1998}).} The cognitive architecture posited by GWT thus contrasts with a \emph{pairwise} architecture, where modules communicate with one another directly rather than pooling their information into a shared space \citep{goyal2021}.

When it comes to functional claims, GWT posits three important aspects of global information processing. First, there is \emph{uptake}, the process that determines which information coming from the modules will enter the global workspace. Second, there is \emph{broadcast}, the process whereby information in the global workspace is sent to a wide range of modules for use in further parallel processing. Finally, there is \emph{processing}, wherein the global workspace integrates and manipulates information it has taken up to shape the trajectory of conscious experience.

GWT distinguishes between two types of processing: parallel processing in individual modules and serial processing in the global workspace. Because it is parallel, processing in individual modules can handle effectively unlimited amounts of information. The global workspace, on the other hand, can process a limited amount of information at any given time. The limited storage capacity of the global workspace creates a bottleneck for information processing which drives the need for selective attention during the uptake phase.

Though it is beyond the scope of our discussion here to present GWT in complete detail, we pause to expand briefly on each of the components of the function of the global workspace and describe some of the psychological phenomena GWT is particularly well suited to explain.

\subsection{Uptake}

At any given time, parallel processing in the modules is generating a large amount of information which could potentially enter the global workspace. \citet{baars1997b} emphasizes three types of information: outer senses, inner senses (including visual imagery, inner speech, and dreams), and ideas. Because the global workspace has limited capacity, however, only some of this information can be taken up from the modules at any given time. There is therefore a kind of competition for information to achieve uptake into the global workspace and to remain there once it does. 

One important example of competition for uptake is binocular rivalry, one instance of a more general class of ‘two-channel experiments’.  In binocular rivalry, a subject is shown a different image in each of their eyes. The subject’s conscious experience flips back and forth between the two images \citep{moreno-bote2011}. Crucially, the subject cannot have simultaneous conscious experiences of inconsistent visual information. This suggests that there is a bottleneck on conscious experience. Consciousness selects among competing visual inputs to generate a coherent narrative about the world.

\citet{baars1988} suggests a number of ways in which this competition might be implemented in the mind. First, informational signals coming from different modules might have different levels of activation, and activation might be increased when different modules work together to boost a particular signal (\citet{baars1988}; p. 95). However, Baars holds that activation alone is not sufficient to get information stably into the global workspace: in addition to high levels of activation, information in the global workspace must receive positive feedback from the modules to which it is broadcast (\citet{baars1988}; p. 205).\footnote{This aspect of Baars’s theory is motivated by the existence of \emph{redundancy} effects, wherein information which is at first consciously accessible fades from consciousness as a subject becomes habituated to it (\citet{baars1997b}; 93). Positive feedback from receiving modules is supposed to capture the idea that a given piece of information in the global workspace contains new information useful to the cognitive system.}

Ultimately, conscious uptake is controlled by attention \citep{posner1994}. This includes both conscious and unconscious attentional processes. Attention boosts information into the global workspace and helps to keep it there (\citet{baars1988}; 308). For example,  unconscious attentional processes explain why, even if we are closely attending to one thing, other information can “break through” into consciousness if it is sufficiently important: ``Absorption does not protect us from high-priority signals, even those that begin unconsciously.'' ( \citet{baars1997b}; 107).

The role of attention in uptake is illustrated by other `two-channel' variants of binocular rivalry: listening to two conversations at once, reading alternating words of text, or seeing two sports games superimposed on the same screen (\citet{baars1997b}; 23-24). In these cases, unconscious information from the second channel will break through if it is sufficiently important, emotional, or relevant. Information in the unconscious channel can also unconsciously influence interpretation of ambiguous words (like \emph{bank}) in the conscious channel (\citet{baars1997b}; 27).  

The \emph{attentional blink} is another empirical finding that showcases the way in which attention shapes conscious experience. In the attentional blink, a pair of stimuli are shown in rapid succession. If the first stimulus is attended to, and if the second stimulus is shown very quickly after the initial one, the second stimulus is not consciously experienced. But if the subject was not directing their attention at the first stimulus, the second stimulus is experienced consciously \citep{raymond1992, dehaene2017}.

Since Baars, neuroscientists have studied in detail the mechanisms by which humans implement the global workspace architecture. A key concept in this area is \emph{ignition}, a specific hypothesis about how uptake is achieved. According to global neuronal workspace theory, the main development of GWT in neuroscience, ``ignition is characterized by the sudden, coherent, and exclusive activation of a subset of workspace neurons coding for the current conscious content, with the remainder of the workspace neurons being inhibited'' (\citet{mashour2020}; 777). In other words, ignition is a non-linear process whereby a single neural representation suddenly comes to dominate the neuronal workspace, suppressing all competing patterns of activation.

\subsection{Broadcast}

The most important function of the global workspace is to serve as a central repository of information available to the cognitive architecture’s various parallel processors. In addition to taking up information from these processors, then, the global workspace must also transmit information back to them. We refer to this function of the global workspace as \emph{broadcast}.

As we have seen, \citet{baars1988} holds that there is an intimate connection between uptake and broadcast: it is only through broadcasting to a coalition of processors and receiving positive activation signals from them that a representation can remain stably in the global workspace. 

In addition to broadcast from the global workspace to processors that handle functions like perception and memory, broadcast from the global workspace also drives intentional action. Building on \citet{james1890}'s ideomotor theory of action, \citet{baars1997b} (p. 138) suggests that when a goal is conscious, it will automatically begin to recruit unconscious ``effectors'' to promote the goal and cause action, unless inhibited by the conscious representation of a conflicting goal. The process of broadcast from the global workspace is thus the means by which serial processing of perceptual and other inputs leads to coordinated agency.

\subsection{Processing}

According to GWT, the global workspace is not simply a passive repository of information available to various cognitive modules; it plays an active role in processing information. Processing in the global workspace is notable for its generality. As Dehaene et al. put it, ``once we are conscious of an item, we can readily perform a large variety of operations on it, including evaluation, memorization, action guidance, and verbal report'' (\citet{dehaene1998}; 14529). Indeed, there are some kinds of processing that can only occur in the global workspace. For example, \citet{baars1997a} (p. 17)  suggests that conscious processing in the global workspace is required to construct concepts out of two-word compounds like \emph{potato soup}. This is illustrated experimentally by priming effects. Unconsciously seeing the word \emph{dog} makes it easier to identify the word \emph{puppy} consciously seconds later. But priming effects are not observed with compound words, like \emph{potato soup} or \emph{honey cake}: the global workspace, and consciousness, is required to integrate the meaning of the two words. Moreover, unconscious priming effects do not influence actions: again, the global workspace is required.

A more general aspect of information processing in the global workspace concerns the demand for consistency. The global workspace seeks to organize disparate information into a single coherent model of the world. One illustration of this demand is visual illusions like the impossible trident and Escher’s infinite staircases. When we visually attend to these images, the global workspace tries to create a single coherent image, even though there isn’t one available. In the impossible trident, tracing the trident across the scene will result in alternating visual images, because there is no way to combine the information into one coherent object (\citet{baars1997b}; 88). 

Information processing in the global workspace is closely connected to some of the core operations of working memory, including the visuospatial sketchpad and phonological loop. The visuospatial sketchpad allows for storage and manipulation of mental images; the phonological loop briefly stores information related to speech, and also allows for its rehearsal \citep{hitch1976}. These processing activities can only trigger when the item of information has been accessed in the global workspace using attention: ``attended memory items can activate subsequent memory states in order to retrieve an association or as part of a cognitive routine when, for example, a mental image is transformed during mental rotation… whereas activity-silent states merely store previously computed states.'' (\citet{mashour2020}; 785).

A final aspect of processing in the global workspace concerns the idea of \emph{effort}. \citet{dehaene1998} suggest that mental effort is related to processing in the global workspace. Many modular perceptual and motor tasks do not involve a phenomenology of mental effort, even when they are complex tasks. By contrast, even simple conscious mental tasks like subtraction do involve effort.

\section{What is Essential to Consciousness? Existing Work}

So far, we’ve surveyed a range of claims that GWT makes about consciousness in humans. But we haven’t said precisely which properties of conscious human systems are necessary and sufficient for consciousness. This is a difficult question. Several authors have noted that it is tricky to say exactly how similar a system needs to be to a human global workspace in order to count as conscious \citep{carruthers2019, birch2022, butlin2023}. More permissive conceptions of consciousness, according to which conscious systems might be less like the human brain, will make it easier for AI systems to be conscious, while less permissive conceptions, according to which conscious systems must be more like the human brain, will make it harder.

As we noted above, we think about these questions through a computational and functionalist perspective on consciousness. The idea here is that there is some functional role associated with consciousness, so that all and only systems instantiating this functional role are conscious. More carefully, we can also think about a necessary and sufficient functional role for a particular mental state to be conscious, and then say that a system is conscious when one of its mental states is conscious. 

It is consistent with this methodology that the relevant functional role is not \emph{metaphysically} necessary and sufficient for consciousness. Rather, for us what is relevant is that the functional role in question be \emph{nomically} necessary and sufficient for consciousness. This distinction allows us to sidestep tricky questions about phenomenal zombies — hypothetical beings that are physically like conscious systems but lack conscious experience. Perhaps phenomenal zombies are metaphysically possible, and so for any functional role, it is metaphysically possible to satisfy it without being phenomenally conscious. Nonetheless, there could still be functional roles that perfectly co-vary with consciousness as a matter of physical law. Another way of thinking about this is that the scientific laws of our world may imply that access consciousness (understood in terms of the global workspace) is sufficient for phenomenal consciousness. If so, AI systems with access consciousness would also be phenomenally conscious.

In this section, we present a number of existing characterizations of the conditions an artificial system would need to satisfy to have a global workspace according to GWT. Though they are on the right track, we believe that none of these characterizations get things quite right. We argue for our own set of conditions in section 5.

\citet{butlin2023} is the most influential recent treatment of the question of what conditions are necessary and sufficient for a system to be phenomenally conscious according to GWT. Butlin et al. propose a set of four conditions:

\begin{itemize}
\item[(B1)] Multiple specialized systems capable of operating in parallel (modules).
\item[(B2)] Limited capacity workspace, entailing a bottleneck in information flow and a selective attention mechanism.
\item[(B3)] Global broadcast: availability of information in the workspace to all modules.
\item[(B4)] State-dependent attention, giving rise to the capacity to use the workspace to query modules in succession to perform complex tasks.
\end{itemize}

In connection with condition (B2), Butlin et al. write that,

\begin{quote}
``...the capacity of the workspace must be smaller than the collective capacity of the modules which feed into it. Having a limited capacity workspace enables modules to share information efficiently, in contrast to schemes involving pairwise interactions such as Transformers, which become expensive with scale… The bottleneck also forces the system to learn useful, low-dimensional, multimodal representations… With the bottleneck comes a requirement for an attention mechanism that selects information from the modules for representation in the workspace.'' (2023: 26–27)
\end{quote}

In addition to Butin et al., there are also several recent papers which provide functional characterizations of GWT without explicitly addressing issues of phenomenal consciousness. Since we are understanding GWT as a theory of the conditions which must be satisfied for a system to be phenomenally conscious, however, these characterizations are relevant to our project. 

\citet{vanrullen2021} (p. 3) propose ``a step-by-step attempt at defining necessary and sufficient components for an implementation of the global workspace in an AI system,'' according to which a system must meet the following conditions: 

\begin{itemize}
\item[(VK1)] A number of independent specialized modules, each with its own high-level latent space.
\item[(VK2)] An independent and intermediate shared latent space, trained to perform unsupervised neural translation between the latent spaces from the specialized modules.
\item[(VK3)] The system prioritizes competing inputs to the shared space with attention.
\item[(VK4)] When a specific module is connected to the workspace as a result of attentional selection, its latent space activation vector is copied into the [global latent workspace] and then immediately broadcast into the latent space of all other modules.
\end{itemize}

VanRullen and Kanai explicate the notion of a `high-level latent space' as follows: ``a latent space is a representation layer trained to encode the key elements of an input domain. This information corresponds to high-level conceptual representations such as visual object features, word meaning, chunks of action sequences, etc.'' (2021: 3).

According to \citet{juliani2022}, on the other hand, a system implements a global workspace just in case it contains a central workspace module that:

\begin{itemize}
\item[(J1)] Has the ability to interact with a dynamic set of modules.
\item[(J2)] Has the capacity for selective attention over the modules.
\item[(J3)] Has the capacity to maintain information over time.
\item[(J4)] Has the capacity to manipulate information over time.
\end{itemize}

Juliani et al. understand (J1) as requiring that the central workspace receive inputs from a set of modules that potentially changes over time, remarking that ``while the neural anatomy of the brain is largely fixed, what counts as a `module' within the GWT is dynamically determined based on population activity across multiple brain regions at a given time'' (957). Their (J2) is intended to capture both attentional processes affecting which representations make their way into the central workspace (ignition) and attentional processes affecting which information in the workspace makes its way back to the modules (broadcast).

While a number of themes emerge from these proposed sets of necessary and sufficient conditions, it is clear that they are not equivalent. (J1), for example, demands that the set of modules connected to the global workspace be dynamic, whereas this requirement is not to be found in the proposals of Butlin et al. or VanRullen and Kanai. And (B2) specifies that the global workspace must have a limited capacity, which is a constraint not imposed by VanRullen and Kanai or Juliani et al. A key philosophical problem in approaching the literature on GWT and artificial systems, then, is to assess the plausibility of each condition proposed as a gloss on GWT.

\section{What is Essential to Consciousness? Theoretical Choice Points}

The existing proposals described in the previous section raise a number of broader questions about how to think about consciousness in AI systems in the context of GWT. We highlight some of these questions in this section before arguing for our considered view in the next.

A first question is how to functionally capture GWT’s notion of attention. \citet{butlin2023} suggest that consciousness requires “top-down” attention, where the information in the workspace can control what further information enters the workspace, in addition to “bottom-up” attention, where the strength of the signals from various modules determines which information enters the workspace. In contrast, though VanRullen and Kanai mention this distinction between top-down and bottom-up attention, they do not build it into their conditions on phenomenal consciousness.

A second question concerns the strength of the broadcast condition. Whereas (J1) simply requires that the workspace interacts with some modules, (B3) and (VK4) require that information in the global workspace is broadcast to every single module. Here, one especially relevant sub-question is whether information in the workspace must be sent back to the perception modules, which would enable top-down perceptual processing. 

A third question concerns the status of information processing in the global workspace. VanRullen and Kanai do not require the global workspace to engage in processing at all, whereas Juliani et al. do but remain unspecific about the nature of this processing. Butlin et al. do not clarify how much processing is required for state-dependent attention, so it is not clear where they come down on this issue. One might require that the workspace engage in very specific types of processing familiar from human working memory; for example, the presence of a visuospatial sketchpad and phonological loop. Alternatively, one could follow \citet{juliani2022} in being unspecific about the nature of the required processing, or not require processing to take place within the global workspace at all.

A final question, and one which in our opinion has been insufficiently discussed in the existing literature on GWT and artificial systems, is whether the representations in the global workspace must have specific structural features. Here one especially important idea is that phenomenal consciousness requires that perceptual representations are rich in some way. For example, \citet{rosenthal2005} suggests that phenomenal experiences are representations that occupy a similarity space, and \citet{tye1995}'s PANIC theory requires specific types of non-conceptual contents. Similarly, \citet{carruthers2020} argues that ``phenomenal consciousness is access-conscious nonconceptual content,'' because it involves especially fine-grained content. Baars glosses richness in terms of relative detail: 

\begin{quote}
``Try raising your eyebrows, for example. Did you know which muscles to contract? Now compare this knowledge to \emph{seeing yourself} raising your eyebrows in a mirror. Which task provides more detailed information, \emph{doing it} or \emph{seeing yourself doing it}? It is commonly reported that we have little or no conscious access to the details in action control, while perception is full of rich detail''(1997b; 64). 
\end{quote}

This all raises the possibility that artificial systems could lack phenomenal consciousness if their perceptual systems were insufficiently rich. 

The debates just discussed are related to a bigger-picture question about the appropriate level of functional analysis for theorizing about consciousness. Requiring only high-level functional similarity to humans results in a more permissive account of phenomenal consciousness, whereas requiring lower-level functional similarity results in a more restrictive account. We ourselves tend to think about the functional role of consciousness in terms of high-level concepts related to manipulating information. But a competing approach might focus more on the implementational level (compare \citet{cao2022}). According to this alternative picture, consciousness functionally requires a neural net that implements higher-level functional ideas in a similar way to humans. For example, uptake might require a neural net that displays similar nonlinear ignition patterns to humans. It might also require an unconscious attention system that is trained in a similar way as the human attention system. By contrast, more abstract levels of analysis might allow an uptake system that implements an informational bottleneck in a different way. 

Even in the case of humans, a higher level functional perspective may be more appropriate for thinking about the global workspace. Baars et al. argue that there is no single brain region that implements the global workspace: ``a brain-based GW capacity cannot be limited to only one anatomical hub. Rather, it should be sought in a dynamic and coherent binding capacity — a functional hub — for neural signaling over multiple networks'' (2013; 20). Ultimately, they suggest that the global workspace recruits a wide range of brain regions and can in principle engage with arbitrary neurons in the brain. If there is no particular brain region associated with the global workspace, this may make it more likely that the essential functions of the global workspace could be implemented in a wide range of systems, including artificial ones. 

\section{What is Essential to Consciousness? A Theory}

How do we assess the competing proposals of section 3 and answer the thorny questions of section 4? In this section, we consider two strategies: first, appealing to the ways in which consciousness is advantageous to systems that possess it; second, employing thought experiments about which changes to conscious systems would turn them into systems that are not conscious. We suggest that each strategy tends to favor more permissive, higher-level functional roles.

\subsection{The Value of Consciousness}

A first way of approaching the problem of distilling a set of necessary and sufficient conditions for consciousness from GWT is to consider what the theory says about the usefulness of consciousness. What purpose does consciousness have in systems that possess it? We might find that consciousness is steered towards solving particular kinds of problems, but that only some and not all of the candidate functional roles sketched in the previous two sections are necessary in order to solve those problems. In general, this approach assumes that phenomenal consciousness is causally efficacious, connecting in lawlike ways to the cognitive and even physical behavior of the organisms that possess it. The methodology is to hold that a property is part of the functional role of consciousness according to GWT just in case that property plays a suitably central role in explaining how having a global workspace helps systems solve certain canonical problems. 

What problems does consciousness help conscious systems solve? \citet{baars1997b} describes a number of information processing tasks that consciousness facilitates on the GWT picture; here, we distill them into four. First, consciousness helps to \emph{classify} inputs to central cognition by prioritizing more important information over less important information. Second, consciousness enforces the \emph{coherence} of the information being processed in central cognition. Third, consciousness helps to \emph{coordinate} cognition by recruiting a range of unconscious tools to solve a single problem. Fourth, consciousness allows for \emph{correction} of errors in reasoning. We suggest that when we think about consciousness in terms of serving these purposes, we should favor more permissive approaches to its functional role.

\begin{itemize}
\item[1.]\textbf{Classification.} Any cognitively sophisticated agent faces the problem of deciding which information is important and relevant to its current situation and which is not. Solving this problem is crucial for reducing the computational complexity of reasoning about how to act. As Dehaene et al. put it: 

\begin{quote}``The organization of the brain into computationally specialized subsystems is efficient, but this architecture also raises a specific computational problem: the organism as a whole cannot stick to a diversity of probabilistic interpretations; it must act and therefore cut through the multiple possibilities and decide in favor of a single course of action. Integrating all of the available evidence to converge toward a single decision is a computational requirement that, we contend, must be faced by any animal or autonomous AI system'' (2021; 3).
\end{quote}

By its nature, the information bottleneck induced by the global workspace requires a conscious system to classify some information as more important than other information — it is only the information that makes it from the modules into the global workspace that is broadcast to the entire system and available for further processing. 

\item[2.]\textbf{Coherence.} A related function of the global workspace is to create a coherent narrative about the world on the basis of potentially inconsistent representations. In the process of selecting information for uptake using attention, the global workspace works to produce consistent interpretations. We saw this demand for coherence in cases of binocular rivalry and other two-channel experiences, as well as in the conscious experience of perceptual illusions like the impossible trident. On the present picture, coherent conscious experience is a tool for facilitating effective agency. If our action-guiding representations of the world change dramatically from moment to moment, we will struggle to form and execute effective plans over time.

\item[3.]\textbf{Coordination.} In a cognitive architecture containing parallel processing modules, solving problems may often require coordinating the activity of multiple processors. The global workspace solves this problem by making the outputs of one processor available to the others and by guiding the activity of the processors over time through top-down attention.

\item[4.]\textbf{Correction.} Top-down attention also facilitates error detection and correction. \citet{baars1997b} argues that one function of consciousness is to monitor for errors: ``if we have no conscious access to our own performance, and if some reliable source of information tells us that we are doing quite badly, we tend to accept misleading feedback because we cannot check our own performance directly'' (136). This effect is illustrated by an experiment due to \citet{langer1979}, where a complex coding task was performed by two otherwise identical groups, who were given the labels ``Bosses'' and ``Assistants''. Each group performed the same task, but the ``Assistants'' group was told they were bad at the task. When doing the task consciously, the two groups performed the same: their conscious access enabled error detection. But once the task was automatic, the ``Bosses'' outperformed the ``Assistants,'' because misleading feedback about ``Assistant'' abilities could not be corrected by conscious error detection. 

The ability to monitor for errors is connected again to the nature of agency, because it is also connected to the concept of voluntary control: ``with direct conscious access to our own performance we are much less influenced by misleading labels. These results suggest that three things go together: consciousness of action details, voluntary control over those details, and the ability to monitor and edit the details.'' (\citet{baars1997b}; 136)
\end{itemize} 

In general, we believe that focusing on the practical advantages conferred by consciousness favors more permissive accounts of the conditions which must be satisfied for consciousness. For example, it is difficult to see why the usefulness of consciousness would require representations in the global workspace to be broadcast back to every input module. Or consider the process of ignition as it is understood by the global neuronal workspace hypothesis. During ignition,  information enters the global workspace through a specific non-linear process of neuronal activation. We don’t think that nonlinear neuronal activation of this specific kind is plausibly required for coherent action or efficient information-processing.

In contrast, attention to the practical advantages of having a global workspace does suggest that certain core features of the GWT framework should be in place for conscious experience. For example, it seems to us that classification requires bottom-up attention, while coherence, coordination, and correction require a combination of top-down attention and processing in the workspace. It is worth noting in this connection that, while Butlin et al.’s (B1)-(B4) probably come closest to capturing the capacities identified above, none of the existing proposals we have discussed simultaneously emphasize the need for bottom-up attention, top-down attention, and information processing within the workspace itself.

\subsection{Thought Experiments}

A second way of approaching the question of what it would take for an artificial system to be conscious according to GWT is to consider thought experiments where a candidate condition is removed from a conscious system and we judge whether that system is still conscious.

First, consider the fact that working memory in human beings is limited to just a few items (seven or less). It would be easy to design AI systems with much larger working memories. But the very small size of human working memory is not plausibly essential to consciousness. Imagine that a person’s working memory expanded from seven to ten thousand items. Intuitively, they would not thereby go from being conscious to being unconscious. Instead, they would be \emph{superconscious}, conscious of vastly more.

This thought experiment suggests to us that consciousness does not require that the global workspace have bounded size. In other words, it suggests to us that the restriction to systems with a limited capacity workspace in Butlin et al.’s (B2) is unwarranted. In this connection, it is worth pointing out that efficient information sharing, useful multimodal representations, and attention could all be implemented in a system with an arbitrarily large global workspace.

Second, consider whether having a dynamic set of modules is plausibly a necessary condition on phenomenal consciousness according to GWT. Start with a conscious system with a global workspace dynamically connected at different times to different subsets of a set S of modules. Now imagine that we change the system so that it is always connected to each module in S, but some of the connections at any given time are not active. Could simply creating the extra connections turn a conscious system into one that lacks consciousness? Would the light of conscious inner life suddenly extinguish when the last connection is installed? We think not. This thought experiment suggests to us that consciousness does not require a dynamic set of modules, as in Juliani et al.’s (J1).\footnote{ It might be protested that Juliani et al.’s conception of a dynamic set of modules is one on which the modules that exist at one time might not exist at another later time, not one on which the same modules always exist but might or might not be connected to the global workspace at any given time. But similar considerations suggest that consciousness does not require the  set of modules to  be dynamic in this second sense, either. Take a conscious  system with a global  workspace connected at time t  to a set S of modules. If we constrain the system so that no modules can be added or removed from S, will it cease to be conscious? Intuitively, no — at most, constraining the system in this way would limit the range of conscious experiences it could have.}

Third, consider whether consciousness requires that information represented in the global workspace must be broadcast recurrently back to every module connected to it. Start with a conscious system with a global workspace which both receives information from and broadcasts information to each of its input modules. Now imagine that we add a module to the system which has a one-way connection to the global workspace: it can send information to the global workspace, but not receive information back. Would the addition of this module turn a conscious system into an unconscious one? Intuitively, we think the answer to this question is no. This thought experiment suggests that conditions like Butlin et al.’s (B3) and VanRullen and Kanai’s (VK4) are unnecessarily strong.

Fourth, consider the richness of perceptual experience. \citet{baars2013}  observes that not only visual but also conceptual representations can be conscious. He describes these conceptual conscious experiences as `fringe' or `feeling of knowing' consciousness: ``Feelings of knowing are not imprecise in their underlying contents. They are simply subjectively vaguer than the sight of a coffee cup\dots lacking clear figure-ground contrast, differentiated details and sharp temporal boundaries. However, concepts, judgments, and semantic knowledge can be complex, precise, and accurate.'' (\citet{baars2013}; 15). Now imagine a human being who lost their visual consciousness, but retained their fringe representations. Would this person no longer be conscious? We think they would still be conscious (compare \citet{chalmers2023}). This suggests that rich perceptual inputs should not be construed as necessary for phenomenal consciousness according to GWT.

Finally, consider the plausibility of conditions tied to specific deep-learning architectural assumptions, like VanRullen and Kanai’s (VK2) and (VK4). (VK2) requires that the global workspace have been trained in a specific way, while (VK4) requires that information stored in the modules and global workspace be representable as an activation vector, and also that movement of information from a module to the workspace be interpretable as a copy function on activation vectors. We think it is easy to imagine a conscious system which does not meet these conditions. Consider, for example, a normal adult human. It is not clear to us that the global workspace of a human brain has been trained to perform unsupervised neural translation between different latent spaces because it is not clear to us that the global workspace of a human brain has been trained at all.\footnote{If you think that the plasticity involved in normal neurological development might count as a kind of training in the relevant sense, consider an intrinsic duplicate of yourself which is produced instantaneously out of inanimate components by a sophisticated machine. The brain of such a duplicate would have never undergone training of any kind, but it would be conscious.} Similarly, to copy the activation vector from one space to another in a literal sense, there must be a neuron in the latter space corresponding functionally to each neuron in the former space. If the global neuronal workspace hypothesis is true and the human global workspace is realized by a specific network of long-range neurons, there is little reason to believe that this network is isomorphic in the required way to the brain regions from which it receives inputs. To put it another way, it seems to us more plausible that when information moves from brain a module into the global neuronal workspace, what happens is that a pattern of activation with content C in the module causally produces a distinct pattern of activation with content C in the global neuronal workspace, where this shared content need not be explained by positing structurally identical patterns of neural activity. This suggests that VanRullen and Kanai’s (VK2) and (VK4) are too specific to be plausible.\footnote{Since VanRullen and Kanai’s explicit goal is to articulate a set of necessary and sufficient conditions for implementing a global workspace \emph{in artificial systems}, the fact that their account makes implausible predictions about biological systems might be thought not to constitute a serious objection. Here we would like to make two points. First, given that one of the main motivations for GWT is that it understands consciousness in terms of the high-level functional organization of information processing, it would be odd for the conditions for implementing a global workspace to differ between biological and nonbiological systems. Second, we see no reason why all artificial systems implementing a global workspace would need to share the deep-learning architectural assumptions built into (VK1)-(VK4). Consider, for example, a neuron-by-neuron simulation of a human brain. If biological brains constitute a problem for  (VK1)-(VK4), then simulated brains do, as well.}

We have rejected a number of candidate conditions on consciousness on the basis of thought experiments. It is worth noting at this point that we do not think the same method can be used to challenge other more foundational aspects of the GWT picture. For example, if we imagine a conscious system consisting of a set of parallel processing modules and a global workspace and then remove the parallel processing modules, we have no strong intuition that the resulting system will be conscious.\footnote{We have weaker intuitions about whether starting with a conscious system and then removing its capacity for top-down attention or its workspace’s ability to promote coherent representations would result in a system that is not conscious. There might be room for reasonable disagreement on these points, and for this reason some might hesitate to follow us in building these features into the conditions for conscious experience. This issue is not central for our dialectical purposes, however, since our claim is that it is possible to imagine a simple language agent architecture that meets even our more demanding conditions.}

Combining considerations from the usefulness of consciousness with the results of our thought experiments, here is our considered view of the necessary and sufficient conditions for conscious experience according to GWT. A system is phenomenally conscious just in case:

\begin{itemize}

\item[(1)] It contains a set of parallel processing modules.
\item[(2)] These modules generate representations that compete for entry through an information bottleneck into a workspace module, where the outcome of this competition is influenced both by the activity of the parallel processing modules (bottom-up attention) and by the state of the workspace module (top-down attention).
\item[(3)] The workspace maintains and manipulates these representations, including in ways that improve synchronic and diachronic coherence.
\item[(4)] The workspace broadcasts the resulting representations back to sufficiently many of the system’s modules.\footnote{Correspondingly, according to our view a representation is phenomenally conscious just in case it is being manipulated by the workspace of a phenomenally conscious system.} 
\end{itemize}

\section{Language Agents}

In the rest of the paper, we’ll apply (1)-(4)  to a particular type of AI system: the language agent. Language agents are created by embedding an LLM into a functional architecture with the structure of an agent that acts predictably according to the laws of folk psychology.  The LLM performs all of the relevant information processing of the system, while the functional architecture ensures that this information processing produces coherent agentic behavior. Below, we’ll focus on the language agents developed by Park et al. (2023) as a case study. But there are numerous other examples of language agents, including AutoGPT\footnote{Project available at \url{https://github.com/Significant-Gravitas/Auto-GPT}.}, BabyAGI\footnote{ Project available at \url{https://github.com/yoheinakajima/babyagi}.}, Voyager\footnote{See \citet{wang2023}.}, SPRING\footnote{See \citet{wu2023}.}, and others\footnote{Perhaps the most successful recent agentic application of language models is Devin, billed as the ``first AI software engineer'' (\url{https://www.cognition-labs.com/introducing-devin}). Another recent example of a step towards language agents is Ghost in the Minecraft, where LLMs learn to navigate the game Minecraft \citep{zhu2023}. Mind2Web is a framework for building web agents \citep{deng2024}. A longer list of existing LLM agents can be found here: \url{https://github.com/e2b-dev/awesome-ai-agents}. ChatDev (\url{https://github.com/OpenBMB/ChatDev}) is another multi-agent environment with some similar features to the Park et al. Generative Agents framework.  For further scaffolding techniques that increase the agency of LLMs, see: Tree of Thoughts \citep{yao2024}, LLM+P \citep{liu2023}, GPT-engineer, and RecurrentGPT. In a similar vein, \citet{zhang2024} develop AgentOptimizer, a framework for training language agents without modifying the weights of their underlying language models. For benchmarks measuring the agency of LLMs, with discussion of applications for language agents, see AgentBench \citep{liu2023} and API-bank \citep{li2023}.}. 

Language agents record and store their beliefs, desires, and plans, as well as their perceptual observations, in natural language. The functional architecture with which a language agent is programmed specifies how these sentences recording beliefs, desires, plans, and observations are fed into the LLM as it considers how the agent will act. Indeed, it is the roles assigned to different stored sentences by the architecture of the language agent which make it the case that they count as the agent’s beliefs, desires, and so forth. In this respect, the structure of a language agent mirrors the structure of the human mind according to cognitive theories which posit a distinction between the content of a representation and its functional role. What makes a representation with the content I am eating a desire rather than a belief in the human mind, according to such theories, is the way in which it enters into an agent’s broader cognitive economy, and especially action planning.\footnote{See for example \citet{fodor1987}.}

Cognition requires not only moving and storing information, but also processing it. The information-processing role in a language agent is played by its LLM. Different cognitive tasks within the language agent may rely on the LLM in different ways. For example, the route from perception to memory might require the LLM to summarize a text description of the agent’s observations to highlight those aspects worth recording, while planning action might require the LLM to reason about what course of action would be rational given the agent’s beliefs and desires. While the language agents in which we are most interested rely more or less exclusively on a single LLM for cognitive processing, for our purposes it would not matter if they relied on some combination of different LLMs, or even if some of their cognitive capacities were realized by hand-coded algorithms.
 
For concreteness, let us return to the language agents developed by \citet{park2023}. Park et al.’s agents live in a text-based simulation called ``Smallville''. They observe and interact with their simulated environment and each other via text descriptions of what they see and how they choose to act. Each agent’s perceptual observations are stored in a text file called the memory stream along with other beliefs, including those comprising their text backstory, which specifies their long-term goals and relationships with other agents. At the end of each day, every agent in Smallville calls on the LLM (in this case, gpt3.5-turbo) to generate a plan for the next day in light of the contents of their \emph{memory stream}. These plans flexibly shape how an agent acts the following day.

\begin{figure}
    \centering
\includegraphics[width=0.9\textwidth]{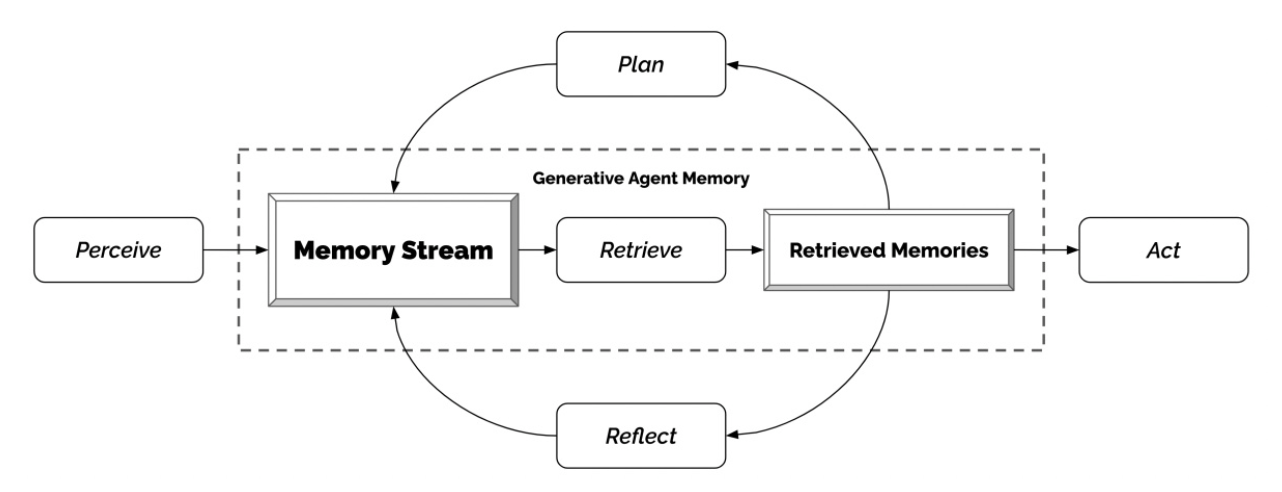}
    \label{fig:agent}
    \caption{The architecture of Park et al.’s language agents. Reproduced from \citet{park2023}.}
\end{figure}

For our purposes in what follows, it will be important to discuss three aspects of Park et al.’s agents in more detail. First, there is the memory stream. In Park et al.’s agents, the memory stream is implicated in all important cognitive processes. It is where perceptions are recorded, as well as where non-perceptual beliefs, desires, and plans are stored. Each entry in the memory stream comes along with a timestamp, and each is assigned an importance score by the LLM as it is recorded. These importance scores play a key role in determining how each entry shapes the future behavior of the agent: entries with higher importance scores are more likely to influence an agent’s other beliefs and plans.

Second, there is the \emph{retrieval function}. Since an agent’s memory stream is unmanageably long, in planning action is it necessary to select an action-relevant subset of entries. This is the role of the retrieval function. Given an input circumstance, the retrieval function produces a list of entries from the memory stream ordered by taking a weighted sum of their importance (as described above), their recency, and their relevance to the input circumstance (this is again calculated by the LLM). In deciding how to act in a given circumstance, each language agent considers only those entries from its memory stream which rank highly enough according to the retrieval function.

Third, there is the process of \emph{reflection}. If the entries in the memory stream were limited to perceptual beliefs and plans, language agents would struggle to function intelligently in light of new information. To solve this problem, Park et al. introduce a function that enables agents to freely draw inferences from their existing beliefs. During reflection, agents draw on the LLM to first pose and then answer a series of questions about how to generalize from their recent experiences. For example, in reflection an agent might pose and then answer questions about their values or the significance of their relationships. The results of reflection are then stored in the memory stream, where they are accessible to the retrieval function.

From the perspective of GWT, Park et al.’s agents have a number of suggestive architectural features. For example, the memory stream maintains representations and interfaces with both the perception and action systems. It also manipulates those representations through reflection, planning, and revising plans in light of new information. There is even an information bottleneck of sorts in the form of the retrieval function imposing constraints on how much of the information in the memory stream can be fed to the underlying LLM.

At the same time, Park et al.’s language agents lack some of the features which we have identified as necessary for consciousness according to GWT. For example, since perception, belief, desire, and planning are all handled by the memory stream, it is not clear that the agents contain a series of parallel processing modules. All information processed by a language agent makes its way into the memory stream, so there is no information bottleneck leading into it. From the perspective of GWT, then, it seems to us that Park et al.’s language agents are not plausibly conscious. This conclusion is in keeping with the general climate of skepticism surrounding claims of consciousness in artificial systems. 

At the same time, however, we wish to challenge this climate of skepticism by suggesting that Park et al.’s language agents can serve as the basis for language agents which would be phenomenally conscious according to GWT — and indeed that the changes necessary to create such agents are technically trivial in the sense that they could be implemented straightforwardly given existing technology.

\section{Language Agents Could Easily Have Conscious Experiences}

We have seen that, while Park et al.’s language agents contain a centralized cognitive module which stores and manipulates information in the manner of a global workspace, it is less clear that they contain a series of input modules to this workspace whose representations must compete for entry into it. In this section, we describe how the architecture of Park et al.’s language agents could be changed to more closely mirror the structure of a conscious system according to GWT. None of the changes we consider will affect how the LLM underlying a language agent is trained or structured. Rather, all of the changes pertain to how that LLM is \emph{scaffolded} to produce an agent. In this way, once we have an LLM that can process information, consciousness will depend on how that information processing is exploited in systematic, lawlike ways by hard-coded rules. 

To begin, note that Park et al.’s language agents in fact have representations of several kinds (beliefs, desires, plans) and engage in a range of forms of cognitive processing including assigning importance scores to memory items, reflection, assigning relevance scores to memory items during retrieval, plan formation and revision, and practical reasoning leading to action. Park et al. make the architectural choice to store all these kinds of representations in the same cognitive workspace, the memory stream, and to have all cognitive processing take as input a series of items from the memory stream. But language agents could also be built with many of the same features arranged into a different architecture.

Imagine, for example, that we modify Park et al.’s architecture in the following ways. First, instead of a single memory stream containing perceptual inputs, other beliefs, desires, and plans, we have a central workspace connected to three further cognitive input modules: a perception module, a belief module, and a desire-and-plan module. Each of these three modules stores representations of the appropriate type. The central workspace can store representations of any type.

Second, imagine that the three input modules perform the following information processing tasks in parallel. The perception module receives observations from the environment and assigns them salience ratings corresponding to how relevant it judges them to be to the agent’s likely future cognition. The belief module assigns each belief an importance score and engages in reflection-driven inference in the same way as Park et al.’s agents. The desire-and-plan module assigns each desire an importance score and engages in a desire-based analog of reflection: generalizing new desires from the agent’s existing list of desires.

Third, imagine that the central workspace plays a number of important information processing roles. It receives perceptual inputs that have been flagged as high salience and decides whether to send them to the belief module for storage. It recursively forms and adds detail to plans on the basis of the agent’s beliefs and desires, working to ensure that the results are coherent. And it chooses how the agent should act on the basis of its beliefs, desires, and plans.

Fourth, imagine that information in the architecture flows along the following paths: in belief formation, from the perception module to the central workspace and then to the belief module; in planning, from the belief and desire-and-plan modules to the central workspace and then back to the desire-and-plan module; in action, from the belief and desire-and-plan modules to the central workspace and then back to the belief and desire-and-plan modules (because actions must be recorded in memory and plans must be revised in light of actions). 

The architecture just described would be no more technically challenging to implement than Park et al.’s architecture. But it would be much closer to the architecture of a phenomenally conscious system according to GWT. In particular, it would contain a series of parallel processing modules and a central workspace that maintains and manipulates representations, including in ways that promote coherence, and broadcasts them back to these modules. The only condition arguably not satisfied would be the idea that representations from the input modules compete through a bottleneck for entry into the central workspace in a way sensitive to both bottom-up and top-down attention.

It is this idea of competition for entry into the global workspace which requires us to depart most significantly from Park et al.’s architecture. But even here, we can take inspiration from Park et al.’s definition of the retrieval function. Recall that the retrieval function generates an ordered list of items from the memory stream which are important, recent, and relevant to an agent’s current situation. We can turn the idea of an ordered list of this kind into something which looks more like competition for entry into the global workspace as follows. Imagine that we define a \emph{competition} function taking as input an ordered triple consisting of the contents of the perception module, the contents of the belief module, and the contents of the desire-and-plan module and yielding as output a set of representations of limited size (for concreteness, suppose we choose 50). Imagine that this function’s output always includes the most salient items from the perception module (for concreteness, suppose there are ten of these), and that the remaining 40 spots are filled by ranking the contents of the belief and desire-and-plan modules according to a weighted combination of their importance, relevance to the agent’s current situation, and recency, as in Park et al.’s retrieval function. Finally, suppose that this competition function is called periodically, and only the representations which it selects have a chance to enter the central workspace.

Implementing a competition function of this kind brings our architecture closer to a canonical GWT system in two ways. First, it introduces an information bottleneck at the point at which the parallel processing modules feed into the global workspace. Second, it gives functional roles to the GWT ideas of bottom-up attention (in the form of the importance or salience the module assigns to a given piece of information) and of top-down attention (in the form of the relevance assigned to different pieces of information given the agent’s current situation).

\begin{figure}
    \centering
\includegraphics[width=0.9\textwidth]{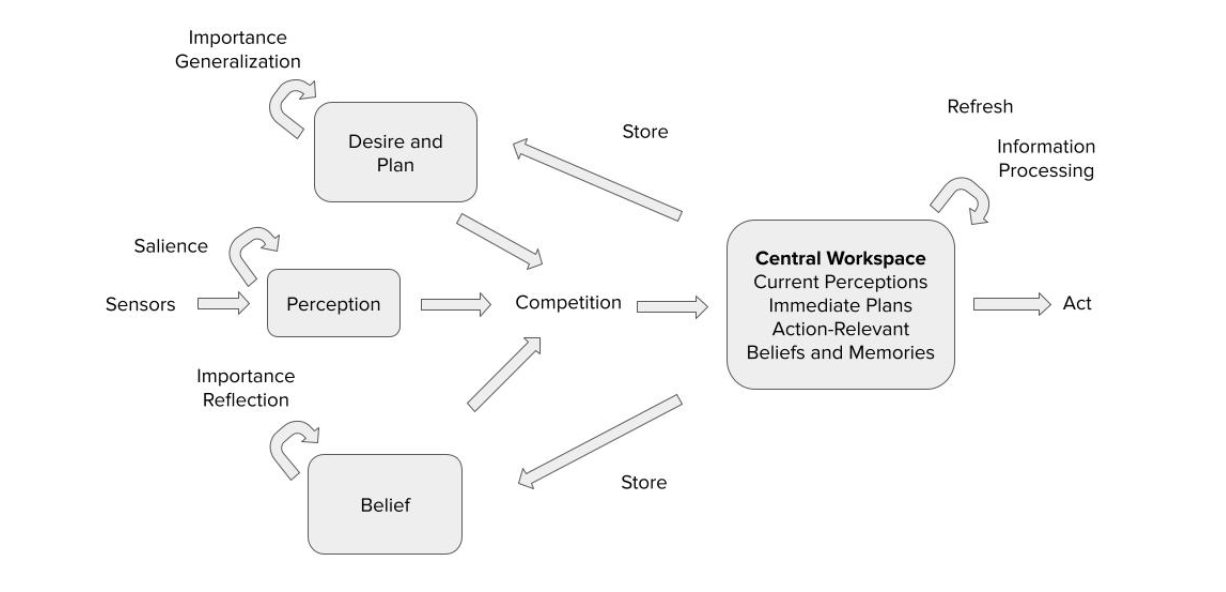}
    \label{fig:architecture}
    \caption{The architecture of a conscious language agent.}
\end{figure}

The language agent architecture we have just described, with its central workspace, three input modules, and competition function, satisfies Section 4’s four conditions for being a conscious system. It contains a series of parallel processing modules: the perception, belief, and desire-and-plan modules. These modules generate representations that compete for entry through an information bottleneck (the competition function) into a central workspace module via a process influenced by both bottom-up and top-down attention. The central workspace module maintains and manipulates the representations in it in ways that promote coherence and broadcasts them back to the belief and desire-and-plan modules. Accordingly, we believe the language agent architecture we have just described is the architecture of a phenomenally conscious artificial system if GWT is correct.

\section{Potential Modifications}

So far, we have argued for a particular view of the necessary and sufficient conditions GWT imposes on phenomenally conscious cognitive systems and described the architecture of a type of language agent which satisfies these conditions. This completes our core case for the near-term possibility of phenomenally conscious artificial systems if GWT is true. 

We are aware, however, that not everyone will agree with us on our characterization of the conditions GWT imposes on phenomenally conscious artificial systems. In this section, we introduce and respond to some further conditions which might be thought relevant in this context.

First, according to both GWT and the global neuronal workspace hypothesis, information in the global workspace does not stay there indefinitely. Instead, it must compete with incoming information from the modules, and, if it loses this competition, it is replaced by that incoming information. To put it slightly differently, one might worry that information in the global workspace needs to be \emph{actively maintained} there. This feature of GWT is not represented in the architecture we described in the previous section.\footnote{Thanks to Rapha\"{e}l Milli\`{e}re for discussion on this point.}

It is not difficult to modify the architecture described above to address this concern. Imagine that each time the competition function is called, the set of representations it generates serves as input to a second function, which we might call the refresh function. The refresh function takes as input the current contents of the central workspace and the output of the competition function. It works by assigning each of the current contents of the central workspace a score based on its importance, relevance to the agent’s current situation, and recency (that is, a score of the same type as those assigned by the competition function) and then discarding any current workspace contents whose score falls below some threshold (for example, items which score lower than the median of the outputs of the competition function). 

Second, \citet{butlin2023} require that the global workspace have a limited capacity. We have argued above that thought experiments suggest that this condition is not necessary for a system to be conscious, but it would be straightforward to implement it in our imagined architecture. We could simply hold that the refresh function ranks the current contents of the workspace and the outputs of the competition function and selects a specific number of top-scoring pieces of information to form the new contents of the central workspace. For example, we could hold that the 50 items provided by the competition function are ranked alongside the current contents of the central workspace, and the top 50 items on this new ranking become the new contents of the central workspace. Setting things up in this way guarantees that the central workspace never contains more than 50 pieces of information.

Third, some may worry that the fact that Park et al.’s language agents function entirely in natural language, with no recognizable perceptual apparatus, prevents them from being phenomenally conscious. Here it is worth emphasizing that while existing language agents are reliant on text-based observation spaces, the technology already exists to implement language agents with richer perceptual modalities. The rise of multimodal language models like GPT-4, which can interpret image as well as text inputs, means that it would not require significant further technological innovation to create language agents with visual perceptual modalities. We can imagine this working in at least two different ways. First, images as perceptual inputs could be translated into text by the perception module using existing captioning technology, and the rest of the agent’s cognition could proceed in natural language. Second, the multimodality of the underlying language model could be used more pervasively throughout the agent’s cognition, so that images could enter the global workspace and be stored in and retrieved from the belief module.

Fourth, it strikes some as significant that the language agents in Park et al.’s paper exist in Smallville, a simulated reality. Might consciousness require appropriate causal connections to the actual world, or a physical body? Again, we think that the technology required to address this kind of worry already exists. Multimodal language models have already been used to control a mobile robotic system — for example, consider Google’s PaLM-E \citep{driess2023}. These agents are embodied and connected to the physical world.

\section{Objections}

Before concluding, we’ll consider two related objections to our approach. First, there is the \emph{small model objection}. This claims that the functional roles we have considered must not be sufficient for consciousness because they can be realized by extremely simple systems. Second, there is the \emph{within/between objection}, which claims that we have conflated two different questions: what distinguishes consciousness from unconsciousness \emph{within} a single organism, and what distinguishes consciousness from unconsciousness \emph{between} organisms.

According to the small model objection, the functional role we have considered cannot be sufficient for consciousness because the distinction between global and local processing can be made in simple neural nets that only have on the order of magnitude of ten neurons \citep{herzog2007}. In particular, we could have two sensory neurons that initially fire in response to distinct inputs. Each such neuron could be a ``perceptual module''. These could feed into a ``global workspace'' neuron system, which could consist of one input neuron that controls ``ignition'' and another that handles ``processing''. This processing neuron could then ``broadcast'' into an ``action module'' neuron that moves a robotic limb, and it could also send information back to the two perceptual modules. Such a five-neuron system might come close to satisfying our functional conditions on consciousness. And we might be able to completely satisfy our functional conditions by adding just a few more neurons to the system to allow for slightly more complex information processing. But it is implausible that a system with so few neurons could be conscious. 

According to the within/between objection, the global workspace theory of consciousness has been used primarily to distinguish conscious from unconscious information within humans: information is conscious to a person when it is represented in their global workspace. But in this paper, we have used this within-organism claim as the basis for a between-organism claim: we have claimed that an organism (or AI system) is conscious when it possesses a global workspace. The advocate of the within/between objection regards this way of repurposing GWT as unjustified.

A first point to make about both of these objections is that they are not especially closely connected to global workspace theory in particular. The small model challenge affects most functional theories of consciousness, since most of these theories embrace functional roles that are elegant enough for a simple system to satisfy.\footnote{One exception to this trend is neurofunctionalism, wherin the functional roles associated with human cognition are so rich that they can only be realized by human brains. See \citet{cao2022} for a recent defense.} This is no coincidence: simplicity is a theoretical virtue, and so a functionalist theory that did not face the small model objection would be less attractive in one important sense. And some version of the within/between objection could be employed against any theory of consciousness which was not developed specifically in the context of artificial systems, since it is always possible to question whether a theory should apply outside the context in which it was originally developed. 

As we understand them, both objections are motivated by the idea that there may be some further necessary condition X on consciousness that is not described by GWT. The proponent of the small model objection takes X to be what is lacked by small models which prevents them from being conscious, while the proponent of the within/between objection takes X to be a feature present in all humans which partly explains why information in the human global workspace is conscious.

We are sympathetic to the small model and within/between objections. At the same time, we think it is important that they not be used merely as a rhetorical strategy to end discussion of the possibility of consciousness in artificial systems. To us, the key questions when we consider any candidate further necessary condition X on consciousness are first, whether X is intuitively plausible and sufficiently motivated from the perspective of the scientific study of consciousness, and second, whether artificial systems like language agents are likely to possess X. We are not aware of any X which is broadly in the spirit of the kind of high-level computational functionalist approach we favor and would rule out consciousness in language agents. For example, it has been suggested to us that X might be the capacity to represent, or the capacity to think, or the capacity for agency.\footnote{Thanks to Dave Chalmers for discussion.} Suppose these choices for X are intuitively plausible and sufficiently motivated by the science of consciousness. Is there any reason to think that language agents like the ones we have described would lack these capacities? We have argued elsewhere that on a wide range of theories of the propositional attitudes, language agents have beliefs and desires \citep{goldstein2023}. It follows that they have the capacity to represent, and (on any plausible theory of thinking) that they have the capacity to think.\footnote{In this context see \citet{chalmers2023}, who ``take[s] thinking to include mental acts such as judging and wondering, as well as dispositional mental states such as believing and desiring'' (25).} While there may be some especially demanding conceptions of agency on which it is not clear that language agents are agents, it seems unlikely to us that such conceptions of agency are sufficiently motivated from the perspective of the scientific study of consciousness. On less demanding conceptions of agency, it seems clear that language agents have the capacity for agency — it’s in the name, after all!\footnote{See, for example, the discussion in \citet{dung2024}.}

Another candidate for X is suggested by Peter Godfrey-Smith’s work on the emergence of consciousness in different organisms.\footnote{See, for example, \citet{godfrey-smith2019}.} Inspired by \citet{merker2005}, he emphasizes the emergence of \emph{self-models} in animals. In one picture, the essence of consciousness is having a \emph{point of view}, and having a point of view is understood in terms of systematically distinguishing changes to the world that are caused by one’s own actions (\emph{reafference}) from external changes. Again, we think that if the language agent architecture we have described does not have this property, it would be technically straightforward to implement it by maintaining a list of the changes in the environment brought about by the agent’s actions. 

It is worth noting that, while we have argued that no choice of X plausibly precludes consciousness in language agents, several of the choices do help with the small model objection. For example, it is not clear that a small model could plausibly be said to think, be an agent, or have a point of view in the relevant sense. For this reason, we take our claim that language agents may be conscious if GWT is correct to be significantly more plausible than similar claims about other systems which have occasionally been made in literature over the past three decades.

\section{Conclusion}

This paper has approached AI consciousness from an architectural perspective. We have identified a series of functional conditions that may be necessary and sufficient for consciousness. Then we have considered the designed architecture of a few AI systems and considered whether this architecture exhibits the relevant functional conditions.

In closing, we’d like briefly to mention another approach to AI consciousness that is still broadly in the spirit of GWT. On this second approach, we test for AI consciousness by focusing directly on the behavioral evidence that cognitive scientists have identified as best explained by a global workspace architecture. For example, we might search for analogues of binocular rivalry, the attentional blink, priming effects, and so on. These phenomena motivate core features of GWT, such as the information bottleneck, modularity, and central processing. With this data in hand, we could say that an AI system is likely to be conscious when it exhibits proper analogues of the behavioral features that have motivated GWT in humans. We hope that future work will critically explore both methodologies, in particular determining whether they make different predictions about any AI systems.

\bibliographystyle{plainnat}
\bibliography{consciousness}

\begin{thebibliography}{49}
\providecommand{\natexlab}[1]{#1}
\providecommand{\url}[1]{\texttt{#1}}
\expandafter\ifx\csname urlstyle\endcsname\relax
  \providecommand{\doi}[1]{doi: #1}\else
  \providecommand{\doi}{doi: \begingroup \urlstyle{rm}\Url}\fi

\bibitem[Baars(1997{\natexlab{a}})]{baars1997a}
B.~J. Baars.
\newblock In the theatre of consciousness. global workspace theory, a rigorous scientific theory of consciousness.
\newblock \emph{Journal of Consciousness Studies}, 4\penalty0 (4):\penalty0 292--309, 1997{\natexlab{a}}.

\bibitem[Baars(1997{\natexlab{b}})]{baars1997b}
B.~J. Baars.
\newblock \emph{In the Theater of Consciousness: The Workspace of the Mind}.
\newblock Oxford University Press, USA, 1997{\natexlab{b}}.

\bibitem[Baars et~al.(2013)Baars, Franklin, and Ramsoy]{baars2013}
B.~J. Baars, S.~Franklin, and T.~Z. Ramsoy.
\newblock Global workspace dynamics: Cortical ``binding and propagation'' enables conscious contents.
\newblock \emph{Frontiers in Psychology}, 4\penalty0 (200):\penalty0 1--22, 2013.

\bibitem[Baars(1988)]{baars1988}
Bernard~J. Baars.
\newblock \emph{A Cognitive Theory of Consciousness}.
\newblock Cambridge University Press, New York, 1988.

\bibitem[Birch(2022)]{birch2022}
J.~Birch.
\newblock The search for invertebrate consciousness.
\newblock \emph{No\^us}, 56:\penalty0 133--153, 2022.

\bibitem[Block(2002)]{block02}
Ned Block.
\newblock Some concepts of consciousness.
\newblock In David~John Chalmers, editor, \emph{Philosophy of Mind: Classical and Contemporary Readings}, pages 206--219. Oxford University Press USA, 2002.

\bibitem[Bramble(2016)]{bramble2016}
B.~Bramble.
\newblock A new defense of hedonism about well-being.
\newblock \emph{Ergo}, 3:\penalty0 85--112, 2016.

\bibitem[Butlin et~al.(2023)Butlin, Long, Elmoznino, Bengio, Birch, Constant, and VanRullen]{butlin2023}
P.~Butlin, R.~Long, E.~Elmoznino, Y.~Bengio, J.~Birch, A.~Constant, and R.~VanRullen.
\newblock Consciousness in artificial intelligence: Insights from the science of consciousness.
\newblock \emph{arXiv preprint arXiv:2308.08708}, 2023.

\bibitem[Cao(2022)]{cao2022}
Rosa Cao.
\newblock Multiple realizability and the spirit of functionalism.
\newblock \emph{Synthese}, 200\penalty0 (6):\penalty0 1--31, 2022.

\bibitem[Carruthers(2019)]{carruthers2019}
P.~Carruthers.
\newblock \emph{Human and Animal Minds: The Consciousness Questions Laid to Rest}.
\newblock Oxford University Press, 2019.

\bibitem[Carruthers(2020)]{carruthers2020}
P.~Carruthers.
\newblock The global workspace.
\newblock The Brains Blog, 2020.
\newblock Available at \url{https://philosophyofbrains.com/2020/01/14/2-the-global-workspace.aspx}.

\bibitem[Chalmers(2023)]{chalmers2023}
D.~J. Chalmers.
\newblock Does thought require sensory grounding? from pure thinkers to large language models.
\newblock \emph{Proceedings and Addresses of the American Philosophical Association}, 97:\penalty0 22--45, 2023.

\bibitem[Dehaene et~al.(1998)Dehaene, Kerszberg, and Changeux]{dehaene1998}
S.~Dehaene, M.~Kerszberg, and J.~P. Changeux.
\newblock A neuronal model of a global workspace in effortful cognitive tasks.
\newblock \emph{Proceedings of the National Academy of Sciences}, 95\penalty0 (24):\penalty0 14529--14534, 1998.

\bibitem[Dehaene et~al.(2017)Dehaene, Lau, and Kouider]{dehaene2017}
S.~Dehaene, H.~Lau, and S.~Kouider.
\newblock What is consciousness, and could machines have it?
\newblock \emph{Science}, 358\penalty0 (6362):\penalty0 486--492, 2017.

\bibitem[Deng et~al.(2024)Deng, Gu, Zheng, Chen, Stevens, Wang, and Su]{deng2024}
X.~Deng, Y.~Gu, B.~Zheng, S.~Chen, S.~Stevens, B.~Wang, and Y.~Su.
\newblock Mind2web: Towards a generalist agent for the web.
\newblock \emph{Advances in Neural Information Processing Systems}, 36, 2024.

\bibitem[Driess et~al.(2023)Driess, Xia, Sajjadi, Lynch, Chowdhery, Ichter, Wahid, Tompson, Vuong, Yu, et~al.]{driess2023}
Danny Driess, Fei Xia, Mehdi~SM Sajjadi, Corey Lynch, Aakanksha Chowdhery, Brian Ichter, Ayzaan Wahid, Jonathan Tompson, Quan Vuong, Tianhe Yu, et~al.
\newblock Palm-e: An embodied multimodal language model.
\newblock \emph{arXiv preprint arXiv:2303.03378}, 2023.

\bibitem[Dung(2024)]{dung2024}
L.~Dung.
\newblock Understanding artificial agency.
\newblock \emph{The Philosophical Quarterly}, 2024.
\newblock Online First.

\bibitem[Fodor(1987)]{fodor1987}
J.~A. Fodor.
\newblock \emph{Psychosemantics}.
\newblock MIT Press, 1987.

\bibitem[Franklin(2003)]{franklin2003}
S.~Franklin.
\newblock A conscious artifact?
\newblock \emph{Journal of Consciousness Studies}, 10:\penalty0 47--66, 2003.

\bibitem[Franklin and Graesser(1999)]{franklin1999}
S.~Franklin and A.~Graesser.
\newblock A software agent model of consciousness.
\newblock \emph{Consciousness and Cognition}, 8\penalty0 (3):\penalty0 285--301, 1999.

\bibitem[Godfrey-Smith(2019)]{godfrey-smith2019}
P.~Godfrey-Smith.
\newblock Evolving across the explanatory gap.
\newblock \emph{Philosophy, Theory, and Practice in Biology}, 11\penalty0 (1):\penalty0 1--24, 2019.

\bibitem[Goldstein and Kirk-Giannini(forthcoming)]{goldstein2023}
S.~Goldstein and C.~D. Kirk-Giannini.
\newblock Ai wellbeing.
\newblock \emph{Asian Journal of Philosophy}, forthcoming.

\bibitem[Goyal et~al.(2021)Goyal, Didolkar, Lamb, Badola, Ke, Rahaman, and Bengio]{goyal2021}
A.~Goyal, A.~Didolkar, A.~Lamb, K.~Badola, N.~R. Ke, N.~Rahaman, and Y.~Bengio.
\newblock Coordination among neural modules through a shared global workspace.
\newblock \emph{arXiv preprint arXiv:2103.01197}, 2021.

\bibitem[Herzog et~al.(2007)Herzog, Esfeld, and Gerstner]{herzog2007}
M.~H. Herzog, M.~Esfeld, and W.~Gerstner.
\newblock Consciousness and the small network argument.
\newblock \emph{Neural Networks}, 20\penalty0 (9):\penalty0 1054--1056, 2007.

\bibitem[Hitch and Baddeley(1976)]{hitch1976}
G.~J. Hitch and A.~D. Baddeley.
\newblock Verbal reasoning and working memory.
\newblock \emph{The Quarterly Journal of Experimental Psychology}, 28\penalty0 (4):\penalty0 603--621, 1976.

\bibitem[James(1890)]{james1890}
W.~James.
\newblock \emph{The Principles of Psychology}.
\newblock Henry Holt and Company, 1890.

\bibitem[Juliani et~al.(2022)Juliani, Kanai, and Sasai]{juliani2022}
A.~Juliani, R.~Kanai, and S.~S. Sasai.
\newblock The perceiver architecture is a functional global workspace.
\newblock In \emph{Proceedings of the Annual Meeting of the Cognitive Science Society}, volume~44, pages 955--961, 2022.

\bibitem[Kagan(1992)]{kagan1992}
S.~Kagan.
\newblock The limits of well-being.
\newblock \emph{Social Philosophy and Policy}, 9\penalty0 (2):\penalty0 169--189, 1992.

\bibitem[Langer and Imber(1979)]{langer1979}
E.~J. Langer and L.~G. Imber.
\newblock When practice makes imperfect: Debilitating effects of overlearning.
\newblock \emph{Journal of Personality and Social Psychology}, 37\penalty0 (11):\penalty0 2014--2024, 1979.

\bibitem[Li et~al.(2023)Li, Song, Yu, Yu, Li, Huang, and Li]{li2023}
M.~Li, F.~Song, B.~Yu, H.~Yu, Z.~Li, F.~Huang, and Y.~Li.
\newblock Api-bank: A comprehensive benchmark for tool-augmented llms.
\newblock \emph{arXiv preprint arXiv:2304.08244}, 2023.

\bibitem[Lin(2021)]{lin2021}
E.~Lin.
\newblock The experience requirement on well-being.
\newblock \emph{Philosophical Studies}, 178:\penalty0 867--886, 2021.

\bibitem[Liu et~al.(2023)Liu, Jiang, Zhang, Liu, Zhang, Biswas, and Stone]{liu2023}
B.~Liu, Y.~Jiang, X.~Zhang, Q.~Liu, S.~Zhang, J.~Biswas, and P.~Stone.
\newblock Llm+p: Empowering large language models with optimal planning proficiency.
\newblock \emph{arXiv preprint arXiv:2304.11477}, 2023.

\bibitem[Mashour et~al.(2020)Mashour, Roelfsema, Changeux, and Dehaene]{mashour2020}
G.~A. Mashour, P.~Roelfsema, J.~P. Changeux, and S.~Dehaene.
\newblock Conscious processing and the global neuronal workspace hypothesis.
\newblock \emph{Neuron}, 105\penalty0 (5):\penalty0 776--798, 2020.

\bibitem[Merker(2005)]{merker2005}
B.~Merker.
\newblock The liabilities of mobility: A selection pressure for the transition to consciousness in animal evolution.
\newblock \emph{Consciousness and Cognition}, 14\penalty0 (1):\penalty0 89--114, 2005.

\bibitem[Mesulam(1998)]{mesulam1998}
M.~M. Mesulam.
\newblock From sensation to cognition.
\newblock \emph{Brain}, 121\penalty0 (6):\penalty0 1013--1052, 1998.

\bibitem[Moreno-Bote et~al.(2011)Moreno-Bote, Knill, and Pouget]{moreno-bote2011}
R.~Moreno-Bote, D.~C. Knill, and A.~Pouget.
\newblock Bayesian sampling in visual perception.
\newblock \emph{Proceedings of the National Academy of Sciences}, 108\penalty0 (30):\penalty0 12491--12496, 2011.

\bibitem[Park et~al.(2023)Park, O'Brien, Cai, Morris, Liang, and Bernstein]{park2023}
Joon~Sung Park, Joseph O'Brien, Carrie~Jun Cai, Meredith~Ringel Morris, Percy Liang, and Michael~S Bernstein.
\newblock Generative agents: Interactive simulacra of human behavior.
\newblock In \emph{Proceedings of the 36th annual acm symposium on user interface software and technology}, pages 1--22, 2023.

\bibitem[Posner(1994)]{posner1994}
M.~I. Posner.
\newblock Attention: The mechanisms of consciousness.
\newblock \emph{Proceedings of the National Academy of Sciences}, 91\penalty0 (16):\penalty0 7398--7403, 1994.

\bibitem[Raymond et~al.(1992)Raymond, Shapiro, and Arnell]{raymond1992}
J.~E. Raymond, K.~L. Shapiro, and K.~M. Arnell.
\newblock Temporary suppression of visual processing in an rsvp task: An attentional blink?
\newblock \emph{Journal of Experimental Psychology: Human Perception and Performance}, 18\penalty0 (3):\penalty0 849--860, 1992.

\bibitem[Rosenthal(2005)]{rosenthal2005}
D.~M. Rosenthal.
\newblock \emph{Consciousness and Mind}.
\newblock Oxford University Press, 2005.

\bibitem[Shanahan(2006)]{shanahan2006}
M.~Shanahan.
\newblock A cognitive architecture that combines internal simulation with a global workspace.
\newblock \emph{Consciousness and Cognition}, 15:\penalty0 433--449, 2006.

\bibitem[Tye(1995)]{tye1995}
M.~Tye.
\newblock \emph{Ten Problems of Consciousness: A Representational Theory of the Phenomenal Mind}.
\newblock MIT Press, 1995.

\bibitem[Van~Gulick(2022)]{van-gulick2022}
R.~Van~Gulick.
\newblock Consciousness.
\newblock The Stanford Encyclopedia of Philosophy (Winter 2022 Edition), 2022.
\newblock Available at \url{https://plato.stanford.edu/archives/win2022/entries/consciousness/}.

\bibitem[VanRullen and Kanai(2021)]{vanrullen2021}
R.~VanRullen and R.~Kanai.
\newblock Deep learning and the global workspace theory.
\newblock \emph{Trends in Neurosciences}, 44\penalty0 (9):\penalty0 692--704, 2021.

\bibitem[Wang et~al.(2023)Wang, Xie, Jiang, Mandlekar, Xiao, Zhu, Fan, and Anandkumar]{wang2023}
G.~Wang, Y.~Xie, Y.~Jiang, A.~Mandlekar, C.~Xiao, Y.~Zhu, L.~Fan, and A.~Anandkumar.
\newblock Voyager: An open-ended embodied agent with large language models.
\newblock \emph{arXiv preprint arXiv:2305.16291}, 2023.

\bibitem[Wu et~al.(2023)Wu, Prabhumoye, Min, Bisk, Salakhutdinov, Azaria, Mitchell, and Li]{wu2023}
Y.~Wu, S.~Prabhumoye, S.~Y. Min, Y.~Bisk, R.~Salakhutdinov, A.~Azaria, T.~Mitchell, and Y.~Li.
\newblock Spring: Gpt-4 out-performs rl algorithms by studying papers and reasoning.
\newblock \emph{arXiv preprint arXiv:2305.15486}, 2023.

\bibitem[Yao et~al.(2024)Yao, Yu, Zhao, Shafran, Griffiths, Cao, and Narasimhan]{yao2024}
S.~Yao, D.~Yu, J.~Zhao, I.~Shafran, T.~Griffiths, Y.~Cao, and K.~Narasimhan.
\newblock Tree of thoughts: Deliberate problem solving with large language models.
\newblock In \emph{Advances in Neural Information Processing Systems}, volume~36, 2024.

\bibitem[Zhang et~al.(2024)Zhang, Zhang, Liu, Song, Wang, Krishna, and Wu]{zhang2024}
S.~Zhang, J.~Zhang, J.~Liu, L.~Song, C.~Wang, R.~Krishna, and Q.~Wu.
\newblock Training language model agents without modifying language models.
\newblock \emph{arXiv preprint arXiv:2402.11359}, 2024.

\bibitem[Zhu et~al.(2023)Zhu, Chen, Tian, Tao, Su, Yang, and Dai]{zhu2023}
X.~Zhu, Y.~Chen, H.~Tian, C.~Tao, W.~Su, C.~Yang, and J.~Dai.
\newblock Ghost in the minecraft: Generally capable agents for open-world environments via large language models with text-based knowledge and memory.
\newblock \emph{arXiv preprint arXiv:2305.17144}, 2023.

\end{thebibliography}
\end{document}